\documentclass[
]{ceurart}

\usepackage[labelfont=bf]{caption}

\begin{document}

\copyrightyear{2021}
\copyrightclause{Copyright for this paper by its authors.
  Use permitted under Creative Commons License Attribution 4.0
  International (CC BY 4.0).}

\conference{Workshop on
Human-in-the-Loop Applied Machine Learning (HITLAML),
September 04-06, 2023 - Belval, Luxembourg}

\title{Beyond original Research Articles Categorization via NLP}
\author[1, 2]{Rosanna Turrisi}[
orcid=0000-0002-0877-7063,
email=rosanna.turrisi@edu.unige.it,
url=https://rosanna-turrisi.webnode.it/,
]
\address[1]{DIBRIS, University of Genova, Genova, 16146, Italy}
\address[2]{Machine Learning Genoa (MaLGa) center, University of Genova, Genova, 16146, Italy}
\begin{abstract}
This work proposes a novel approach to text categorization -- for unknown categories -- in the context of scientific literature, using Natural Language Processing techniques. The study leverages the power of pre-trained language models, specifically SciBERT, to extract meaningful representations of abstracts from the ArXiv dataset. Text categorization is performed using the K-Means algorithm, and the optimal number of clusters is determined based on the Silhouette score. The results demonstrate that the proposed approach captures subject information more effectively than the traditional arXiv labeling system, leading to improved text categorization. The approach offers potential for better navigation and recommendation systems in the rapidly growing landscape of scientific research literature.
\end{abstract}


\maketitle

\section{Introduction}
In the past decade there has been a significant increase in the number of published research papers, creating a need for better tools to navigate through the vast literature. ArXiv\cite{arxiv}, an open-access archive, has emerged as the most popular platform, housing over two million scientific articles in various fields such as physics, mathematics, computer science, biology, finance, statistics, engineering, and economics. Currently, authors manually assign subject categories to their own articles during submission. However, this process is time-consuming and restricts the labels to sector-based categories. Consequently, inter-disciplinary works focusing on similar topics often receive different labels. For example, two articles studying brain cancer -- one using Artificial Intelligence (AI) and the other using statistics -- would be assigned to the categories of \textit{Computer Science} and \textit{Statistics}, respectively, even though they investigate the same phenomenon. On the other hand, the AI-based cancer study would share the category label with a work on Operating Systems, resulting in more challenging literature search and less efficient recommendation systems.\\

Recent advancements in Natural Language Processing (NLP) and the success of pre-trained models \cite{qiu2020pre} have opened up new possibilities for processing text data and performing various tasks such as top modeling \cite{bianchi2020cross, ramage2009topic, grootendorst2022bertopic}, text classification \cite{yang2019xlnet, sun2019fine, liu2018task}, and information retrieval \cite{wang2017knowledge}.

This work leverages NLP to process abstracts of ArXiv papers and classify them into more meaningful and flexible subject categories that go beyond the original labeling. The aim is to create categories providing information about the original subject categories but less restrictive and sector-based. The ultimate goal is to enhance literature search and recommendation systems by providing more accurate and relevant categorization. The proposed approach differs from most studies on text categorization \cite{gonzalez2023landscape, lumbanraja2021abstract} in two main aspects: i) the optimal number of categories ($N$) is unknown, and determining its best value poses a significant challenge; ii) although text categorization is performed in an unsupervised setting, the feature extraction process incorporates knowledge about the original subject labeling resulting in a hybrid approach. 

\paragraph{Main contributions} This study explores four different abstract embeddings based on the SciBERT pre-trained model \cite{beltagy2019SciBERT}. 
Each embedding is used as input for the K-means algorithm, enabling unsupervised text categorization. The optimal number of categories is determined by evaluating the model performance using the Silhouette score on the validation set. Results demonstrate that this approach effectively captures information from the ArXiv subject categories while providing more meaningful text categorization. For instance, it successfully collapses distinct category labels from ArXiv (e.g., \texttt{stat.Th} and \texttt{math.ST}) that correspond to the same subject (e.g., \textit{Statistic Theory}) into a single class category.
The implemented pipeline was developed using the Python programming language. Its code can be accessed on \href{https://github.com/rturrisige/TextClassification}{GitHub}.

\section{Related work}
The automatic classification of research publications is commonly achieved by assigning papers to existing categories within hierarchically-structured vocabularies, such as Medical Subject Headings (MeSH) \cite{MESH}, Physics Subject Headings (PhySH) \cite{PhySH}, and the STW Thesaurus for Economics \cite{STW}. For instance, \cite{mai2018using} introduces three deep learning architectures for article classification, utilizing either the paper title or the full-text as input. Notably, results show that the title-based method performs comparably to the full-text-based approach. Similar investigations are reported in \cite{galke2017using, nishioka2016profiling}, in which the aim is either surpass or achieve results equivalent to the full-text-based approach. This is accomplished by solely utilizing paper titles, which benefit from widespread availability. An intermediate approach is presented in \cite{kandimalla2021large}, where category classification is accomplished using paper abstracts. This strategy enables the utilization of abundant data while capturing more comprehensive and insightful information about paper content. However, these studies thrive only within a very well-defined category framework. Pushing further, various techniques \cite{osborne2012mining, ereteo2011semtagp, salatino2019cso, hitha2021topic, al2022investigating, sharma2023machine} integrate machine learning methods with background knowledge to identify research topics in documents. Despite enhancing automated paper classification quality, these methods rely on ontologies that are time-consuming and require careful planning and expertise. 
This manuscript capitalizes on a domain-specific taxonomy, taking advantage of its immediate accessibility and usability. However, it acknowledges that pre-defined categories might not encompass the entire intricacies of concepts and connections present in a formal ontology. To address this gap, an unsupervised machine learning algorithm is here proposed to extract complex information from paper abstracts and identify research categories that are related to, but distinct from, the pre-existing categories, offering a more refined representation of the research articles.

\section{Dataset}
The ArXiv Dataset \cite{arxiv_dataset} is a rich corpus of approximately two million articles, including author information, title, journal references, ArXiv categories, and abstract. 
 For computational reasons, the subset of articles published in 2023 was selected. To ensure data quality,  duplicated and withdrawn papers were removed from the dataset. Categories with a small number of papers, specifically those containing less than 250 articles, were also excluded. 
 Furthermore, abstracts with fewer than 31 words were not considered in the analysis. This decision was made due to the observation that such short abstracts often contain meaningless or misleading texts, such as sentences indicating revisions or the absence of an abstract for comments. This resulted in a final corpus of 43853 samples and reduced the memory usage
from 240.3MB to 2.3MB.

\subsection{Data analysis}

\paragraph{ArXiv categories.} The selected ArXiv subset comprises 40 unique subject categories (e.g., \textit{Algebric Geometry}) from which 15 macro categories (e.g., \textit{Mathematics}) can be retrieved.

\begin{figure}[t!]
\centering\includegraphics[width=0.9\textwidth]{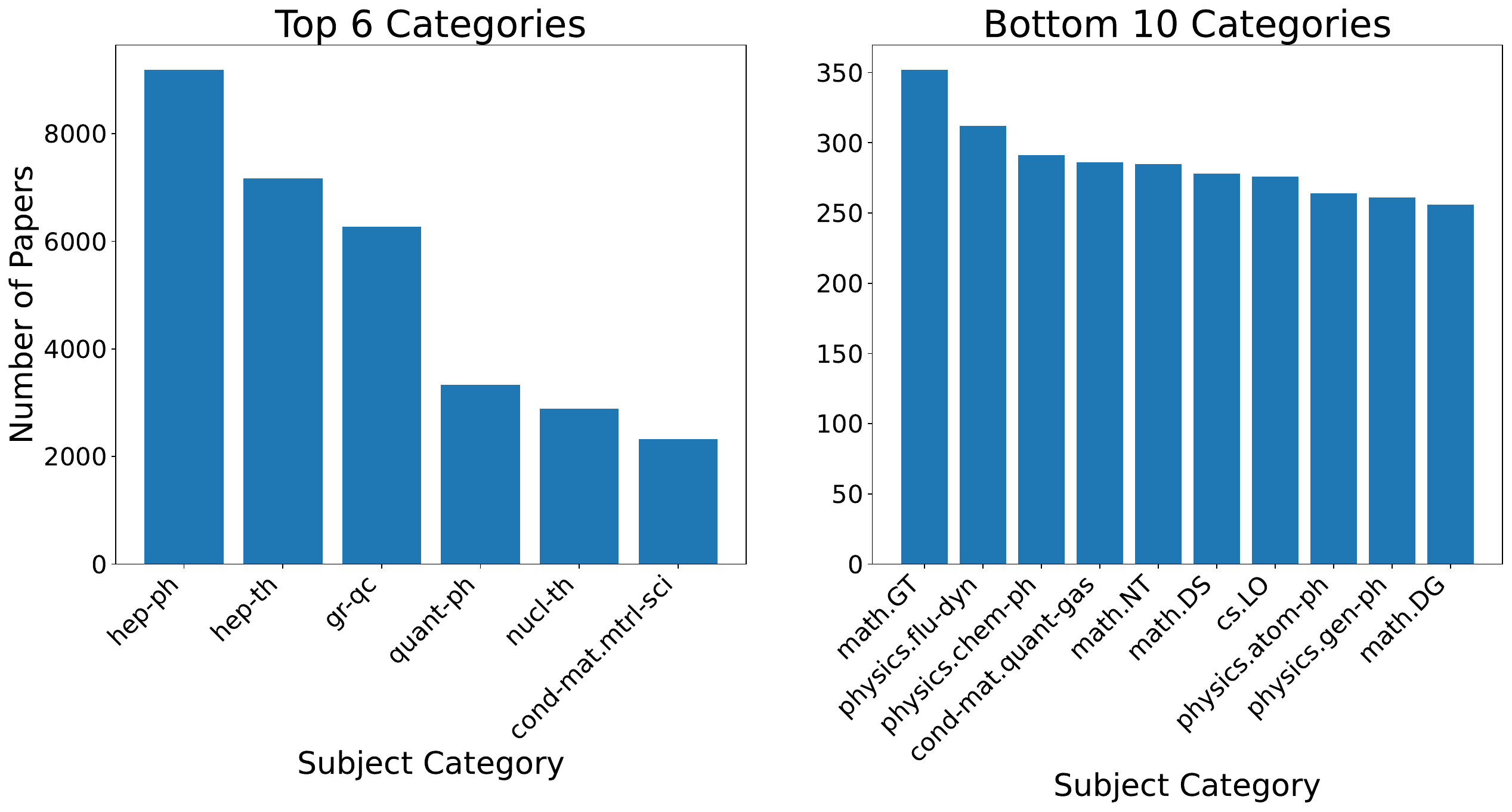}
\caption{Number of papers in the mostly frequent (left) and the less frequent subject categories (right).}
\label{fig:mlcat}
\end{figure}

Fig. \ref{fig:mlcat} (left) shows the number of papers associated with the 6 most frequent categories. Specifically, it displays the number of papers associated to    
\begin{enumerate}
    \item \textit{High Energy Physics - Phenomenology} (\texttt{hep-ph});
    \item \textit{High Energy Physics - Theory} (\texttt{hep-th});
    \item \textit{General Relativity and Quantum Cosmology} (\texttt{gr-qc});
    \item \textit{Quantum Physics} (\texttt{quant-ph});
    \item \textit{Nuclear Theory} (\texttt{nucl-th});
    \item \textit{Materials Science} (\texttt{cond-mat.mtrl-sci}).
\end{enumerate}

    The most frequent category contains more than 9000 papers while the 6th most frequent category includes about 2000 papers.  Fig. \ref{fig:mlcat} (right) reports instead the 10 less frequent categories: 
    \texttt{math.GT} (\textit{Geometric Topology}), 
    \texttt{physics.flu-dyn} (\textit{Fluid Dynamics}), 
    \texttt{physics.chem-ph}( \textit{Chemical Physics}),
    \texttt{cond-mat.quant-gas} (\textit{Quantum Gases}),
    \texttt{math.NT} (\textit{Number Theory}),
    \texttt{math.DS} (\textit{Dynamical Systems}),
    \texttt{cs.LO} (\textit{Logic in Computer Science}),
    \texttt{physics.atom-ph} (\textit{Atomic Physics}),
    \texttt{physics.gen-ph} (\textit{General Physics}),
    \texttt{math.DG} (\textit{Differential Geometry}). As we can see, the number of papers is between 250 and 350 for all of them. 

\begin{figure}[t!]
\centering
\subfloat[]{{\includegraphics[width=0.45\textwidth]{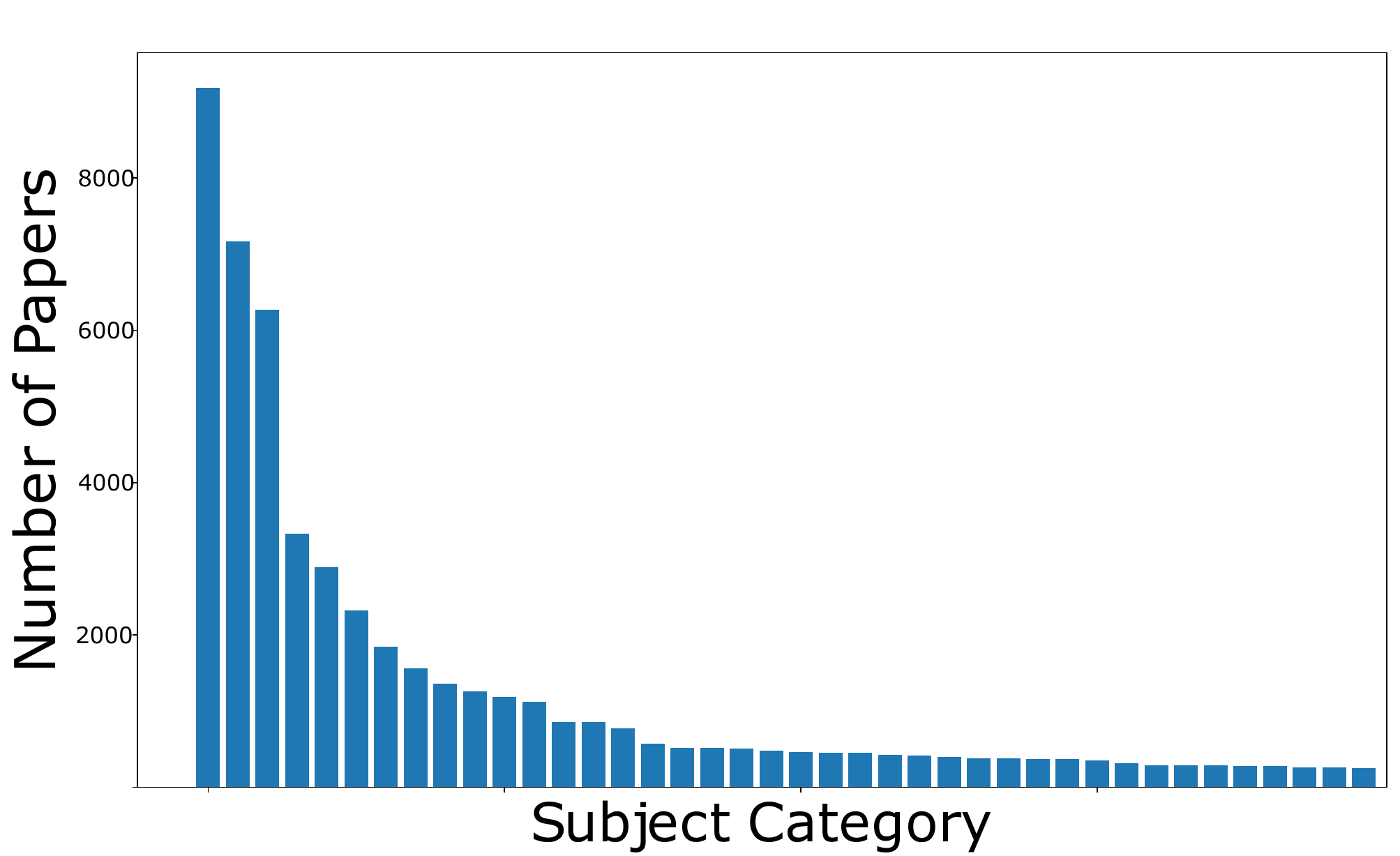}}}
 \qquad
\subfloat[]{{\includegraphics[width=0.41\textwidth]{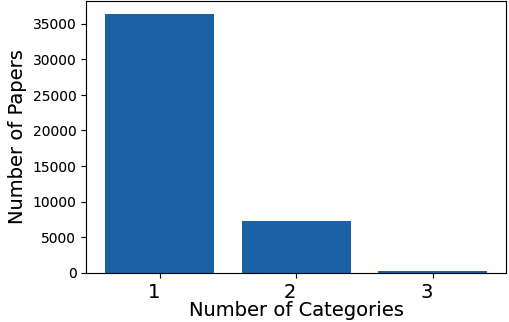}
}}
\caption{ArXiV categories distribution study. (a)  Frequency of all categories. (b) Number of papers with one or more categories associated.}
\label{fig:categories}
\end{figure}

Fig. \ref{fig:categories} (a) provides a visual representation of the number of papers within each category. It is evident from the graph that the distribution of category labels is highly unbalanced and that the large majority of categories include less than 1000 articles. Since some papers may be associated with multiple labels, a histogram illustrating the number of categories associated with each paper is presented in Figure \ref{fig:categories} (b). This histogram provides insights into the distribution of multiple category assignments for individual papers.

 \paragraph{Abstract length.} Fig. \ref{fig1} reports an histogram of abstract length in terms of number of words. As we can see, the abstracts' length typically ranges from 50 to 300 with few exceptions. Table \ref{table1} shows average, standard deviation (Std), minimum (Min), maximum (Max) and first, second and third quartiles (25\%, 55\%, 75\%) of the distribution of abstracts' length.
 This analysis is relevant for the study as the third quartile was used to fix a maximum text length in the tokenization process.

\begin{minipage}[ht!]{0.6\linewidth}
\vspace{15pt}
\centering\includegraphics[width=0.95\textwidth]{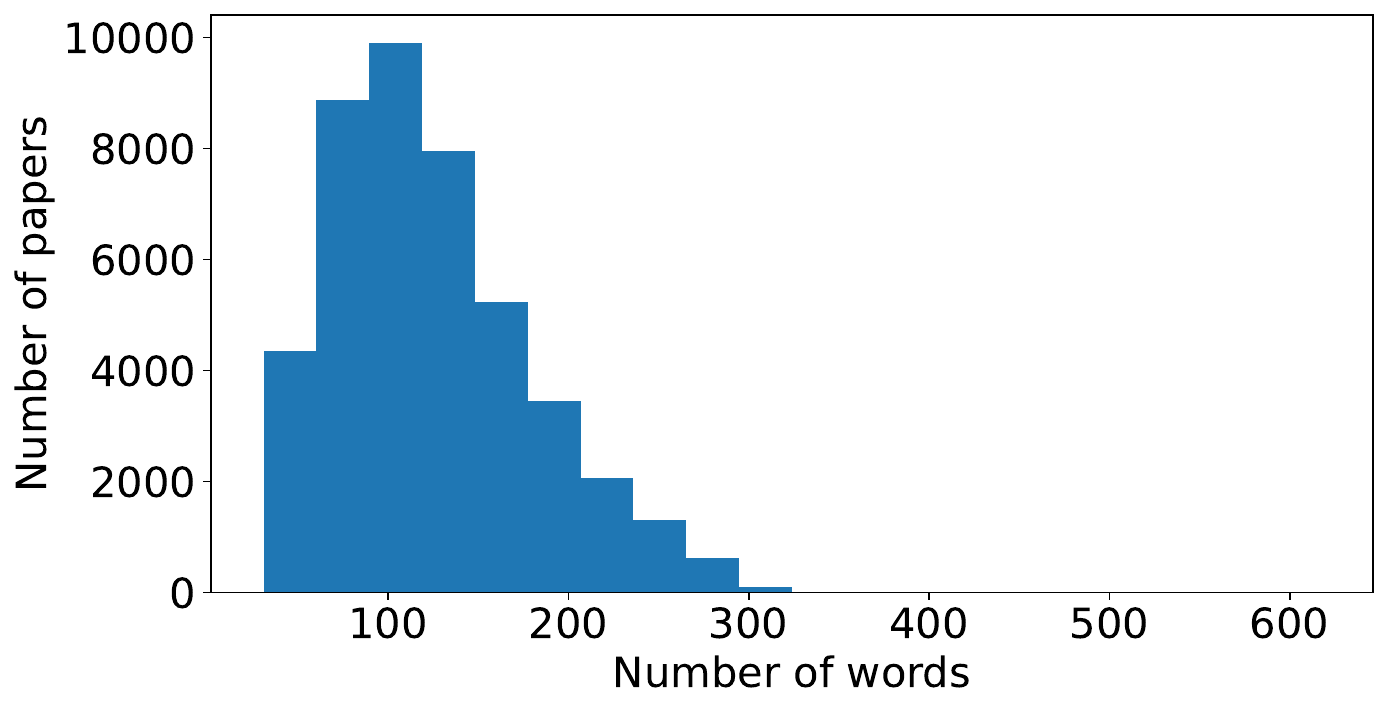}
\captionof{figure}{Abstracts length distribution.}
\label{fig1}
\end{minipage}
\begin{minipage}[ht!]{0.4\linewidth}
\captionof{table}{Abstracts length statistics}
\label{table1}
\begin{tabular}{cc}
    \toprule
N. samples & 43853\\
    \midrule
Mean & 124.5\\
Std & 55.3\\
Min & 31\\
25\% & 83\\
50\% & 115\\
75\% & 157\\
Max & 617\\
    \bottomrule
\end{tabular}
%\end{table}
\end{minipage}
\subsection{Data processing}
Text processing was performed using the \texttt{spaCy} \cite{spacy2} package, specifically using the \texttt{en\_core\_ sci\_lg} component. This component is trained on scientific papers and offers a large vocabulary and 600,000 word vectors. Each abstract in the dataset was first syntactically parsed in order to find linguistic units and their grammar dependencies. Then, the following processing steps were applied: i) the text was converted to lowercase; ii) lemmatization was performed on linguistic units, excluding personal pronouns, to transform them into their base form; iii) punctuation marks and stop words were removed from the text.
Finally, processed data was tokenized using the \texttt{AutoTokenizer} class from the \texttt{Hugging-face/transformers} package \cite{wolf-etal-2020-transformers}.

\section{Experimental setup}

\subsection{Embedding estimation}
Text embedding was performed by relying on the SciBERT pre-trained model \cite{beltagy2019SciBERT}, a language model trained on scientific texts and known to effectively capture domain-specific information. Two different text representations were investigated in this study: i) SciBERT-T, obtained by extracting the last hidden layer of the first token in the input sequence; ii) SciBERT-CLS, obtained by extracting the last hidden layer of the classification token in the input sequence. Both of them provide a 768-dimensional dense vector representation of the input text. 

\paragraph{PCA-SciBERT.}
A feature reduction method was applied to the SciBERT embeddings to address the curse of dimensionality and discard irrelevant features. Specifically, Principal Component Analysis (PCA) was applied separately to the SciBERT-T and SciBERT-CLS representations by retaining the 95\% of the variance in the data. This reduced the SciBERT-T embedding to a 325-dimensional vector, and the SciBERT-CLS embedding to a 122-dimensional vector.

\begin{figure}[t!]
\centering\includegraphics[width=0.75\textwidth]{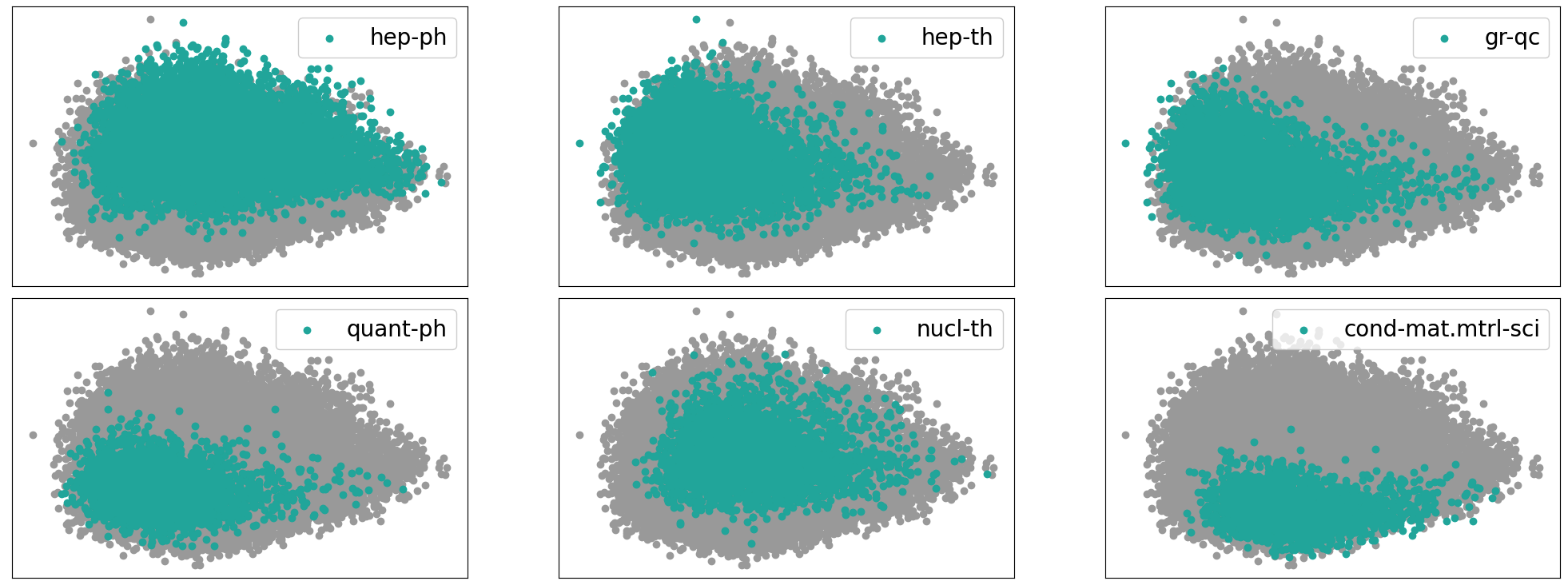}
\caption{2D projection of PCA-SciBERT-T embeddings by using t-SNE. Labelled samples belonging to the same arXiv subject category are shown in green, while the others in grey. The graph reports texts from the 6 most frequent categories.}
\label{fig:embedding_T}

\centering\includegraphics[width=0.75\textwidth]{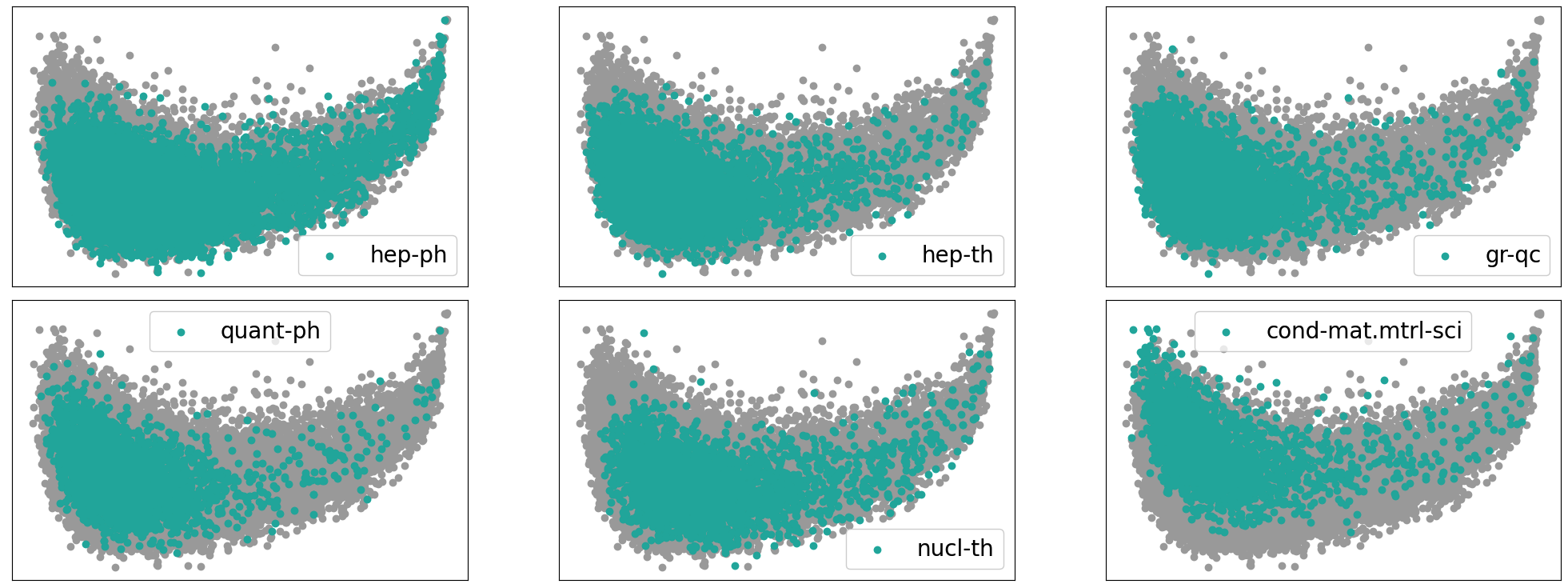}
\caption{2D projection of PCA-SciBERT-CLS embeddings by using t-SNE. Labelled samples belonging the same arXiv subject category are shown in green, while the others in grey. The graph reports texts from the 6 most frequent categories.}
\label{fig:embedding_CLS}
\end{figure}

Fig. \ref{fig:embedding_T} and \ref{fig:embedding_CLS} provide a qualitative visual assessment of the resulting embeddings, highlighting the 6 most frequent subject categories. The t-SNE method \cite{vandermaaten08a} is employed to project the samples into a 2D space. In both figures, samples belonging to the selected ArXiv category are represented as green points, while the remaining categories are displayed in grey. The categories are presented in descending order of frequency.

The observed clustering of the green points in the extracted embeddings indicates a tendency for points with similar category associations to group together. However, it is important to note that the 3 most frequent categories exhibit overlapping clusters, particularly in the PCA-SciBERT-CLS representation. This observation should be interpreted considering two factors: i) a paper abstract can have multiple labels, as illustrated in Figure \ref{fig:categories} (b); and ii) the representation of distant points in a high-dimensional space may not be accurately captured in a 2D projection.

%A qualitative evaluation of the resulting embeddings, based on 2D embedding visualization, has been performed and reported in Supplementary Materials. 
% the study aimed to capture the most relevant information while reducing the computational complexity and potential noise in the data.
\paragraph{FT-SciBERT.} As reported in \cite{beltagy2019SciBERT}, NLP tasks performance is generally improved by Fine-Tuning (FT) the SciBERT model for a small number of epochs. Hence, a FT approach was here explored to enhance text representation. The ArXiv subject category associations were used as labels, with the possibility of multiple labels for each sample. 
To implement the FT, a dense layer with 32 nodes was added on top of the SciBERT model, followed by a classification layer. This additional hidden layer helps incorporate specific prior knowledge about the arXiv dataset into the FT process. While the classification task captures subject information by improving the embedding representation, it should be noted that the embedding will not be a direct representation of the arXiv subject categories due to the presence of the added hidden layer. 

\begin{figure}[t!]
\centering\includegraphics[width=0.75\textwidth]{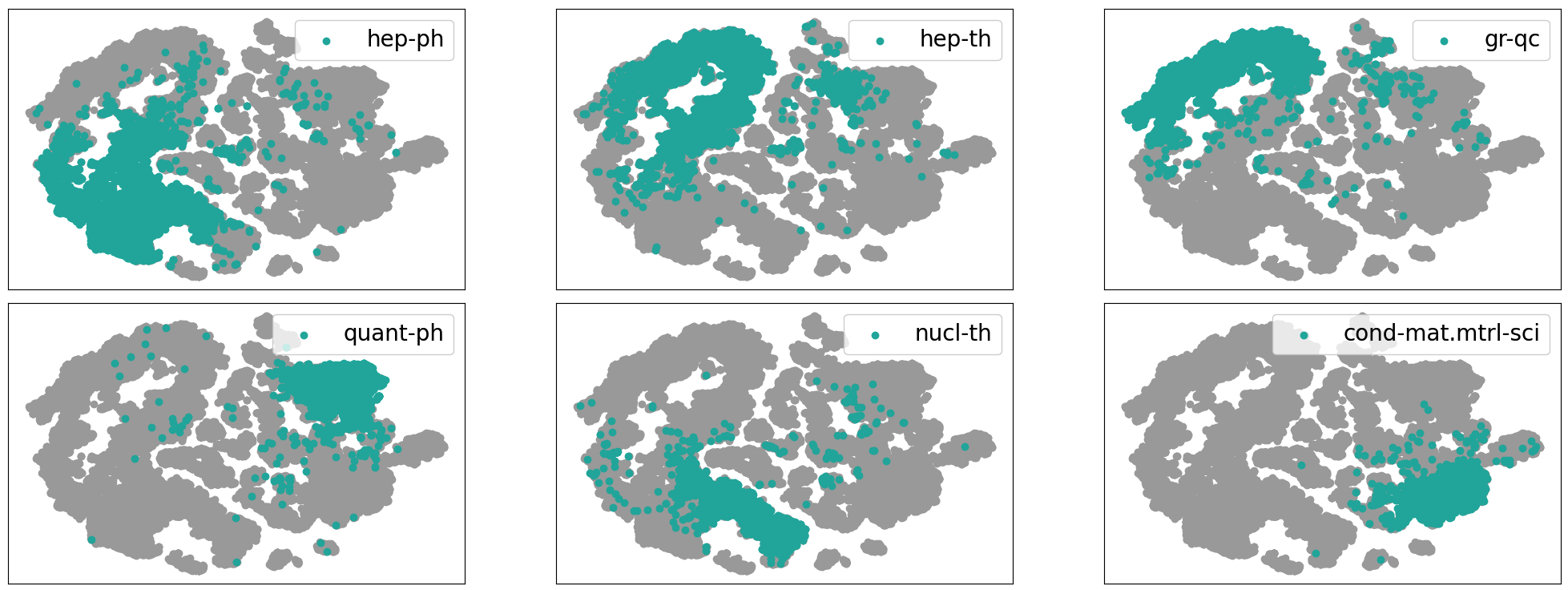}
\caption{2D projection of FT-SciBERT-T embeddings of learning set. Labelled samples belonging the same arXiv subject category are shown in green, while the others in grey. The graph reports texts from the 6 most frequent categories.}
\label{fig:train_embedding_T}

\centering\includegraphics[width=0.75\textwidth]{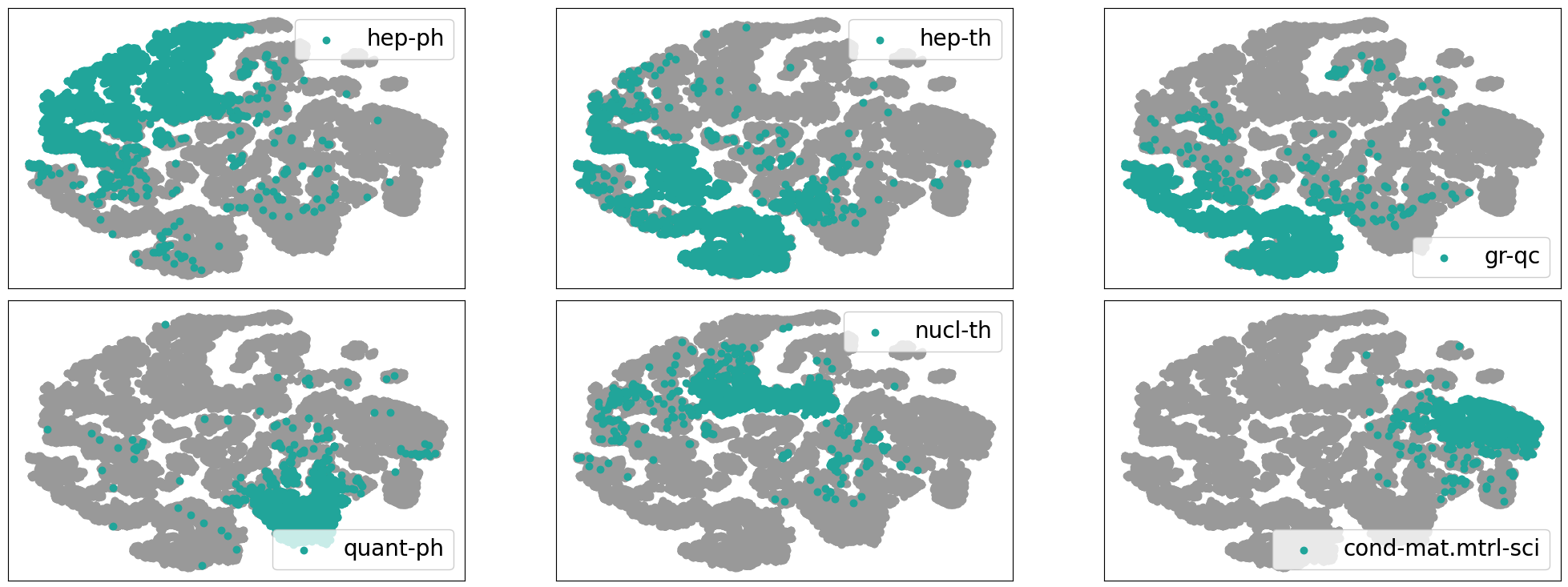}
\caption{2D projection of FT-SciBERT-CLS embeddings of learning set. Labelled samples belonging the same arXiv subject category are shown in green, while the others in grey. The graph reports texts from the 6 most frequent categories.}
\label{fig:train_embedding_CLS}
\end{figure}

 The dataset was split into learning (90\%) and testing (10\%) set. The learning set was further split into training (80\%) and validation (20\%) set. Training was performed by using 32 as batch size, 2e-5 as learning rate, and dropout with a dropping probability of 0.1.  These parameter values were reported in \cite{beltagy2019SciBERT} as the best performing across different datasets and tasks. The maximum number of epochs was set to 4, and early stopping was employed with a patience value of 1, allowing the training process to stop if there was no improvement in performance.\\

Fig. \ref{fig:train_embedding_T} and \ref{fig:train_embedding_CLS} illustrate the embeddings obtained from the learning set using FT-SciBERT-T and FT-SciBERT-CLS, respectively. In all figures, green points represent the 2D projections of the embedding points associated with the 6 most frequent subject categories, while grey points represent other categories. The t-SNE algorithm was used to obtain the 2D projection. As anticipated, fine-tuning the model improved the embedding representation, resulting in more distinct clustering of abstracts with the same arXiv category.
%Similarly, Fig. \ref{fig:test_embedding_T} displays the projection of the testing set embeddings based on FT-SciBERT-T. 

\subsection{Unsupervised text categorization}
Unsupervised text classification was performed using K-Means algorithm for each text representation. The goal was to determine the optimal number of categories, denoted by $N$, for the clustering task. The range of $N$ was set from 2 to 50. To evaluate the clustering performance for different values of $N$, the algorithm was trained on the training set and then evaluated on the validation set using the Silhouette metric. The Silhouette score ranges from -1 to 1, where a score of -1 indicates that points are wrongly assigned to clusters, a score of 0 suggests overlapping clusters, and a score of 1 indicates that points are perfectly assigned to well-separated clusters.
The Silhouette metric was used as a criterion to determine the optimal number of categories. The value of $N$ that yielded the highest Silhouette score on the validation set was taken as the best choice for the number of categories.
 Alternatively, one may adopt the Within-Cluster Sum of Square (WCSS) curve. The WCSS curve measures the sum of squared distances between each point and its assigned cluster centroid. Typically, the curve exhibits an `elbow' shape, and the optimal $N$ corresponds to the point where the curve starts to flatten out significantly. However, in this study, the elbow method was not utilized as the WCSS curve was found to be too smooth, making it challenging to identify a clear elbow point.
 
\section{Results} 
\begin{figure}[t!]
\centering\includegraphics[width=0.8\textwidth]{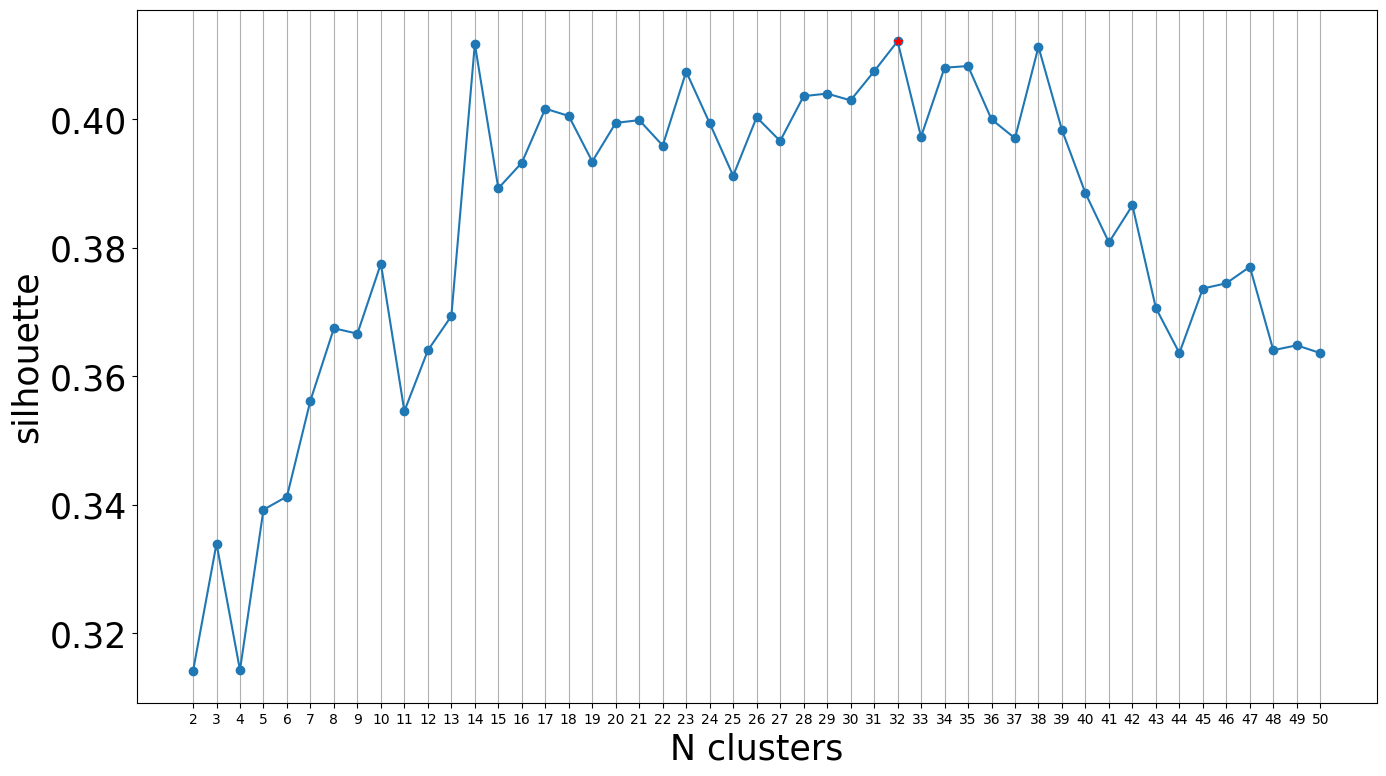}
\caption{Silhouette score of K-Means algorithm applied to FT-SciBERT-CLS for different number of clusters ($N$) ranging from 2 to 50. The red point indicates the highest score achieved at $N=32$.}
\label{fig:sil_score}
\end{figure}
When using PCA-SciBERT representations, the Silhouette scores on the validation set was found to be close to 0 for any value of $N$. This suggests that these embedding points are difficult to associate with non-overlapping clusters, indicating a lack of clear separation between categories.
On the other hand, when using FT embeddings, the K-Means clustering algorithm achieved better results. The Silhouette score for FT-SciBERT-T was 0.36, while FT-SciBERT-CLS achieved a score of 0.41. As the higher Silhouette score for FT-SciBERT-CLS indicated better cluster separation and cohesion, further observations and the evaluation on the testing set were conducted only for the FT-SciBERT-CLS embedding.\\

Fig. \ref{fig:sil_score} presents the evaluation results on the validation set using the Silhouette metric for different numbers of clusters ($N$) ranging from 2 to 50. As expected, the Silhouette score initially increases as $N$ increases, reaching a peak value, and then starts to decline rapidly for $N>38$. The best clustering performance is observed at $N=32$, which falls between the number of macro subject categories (15) and the number of subject categories (40) of ArXiv. This finding suggests that the arXiv categories do not fully capture the underlying structure of the abstract classes.

% This text representation demonstrated the highest model performance on the validation set, suggesting its suitability for the task of unsupervised text classification.

\begin{figure}[t!]
\centering
\subfloat[\centering Green points visualizes articles belonging to same arXiv subject category, while the others are plotted in grey.]{{\includegraphics[width=0.55\textwidth]{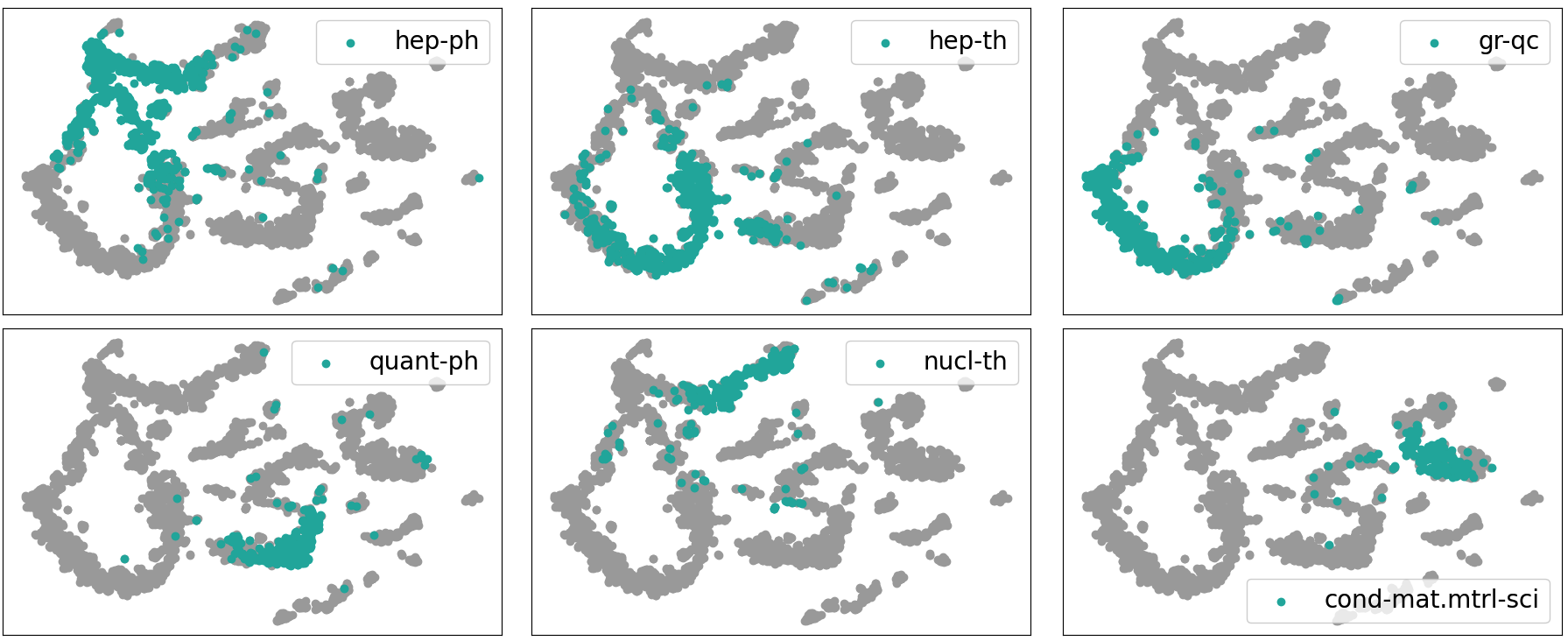}}}\quad
\subfloat[\centering K-Means clusters for $N=32$.]{{\includegraphics[width=0.4\textwidth]{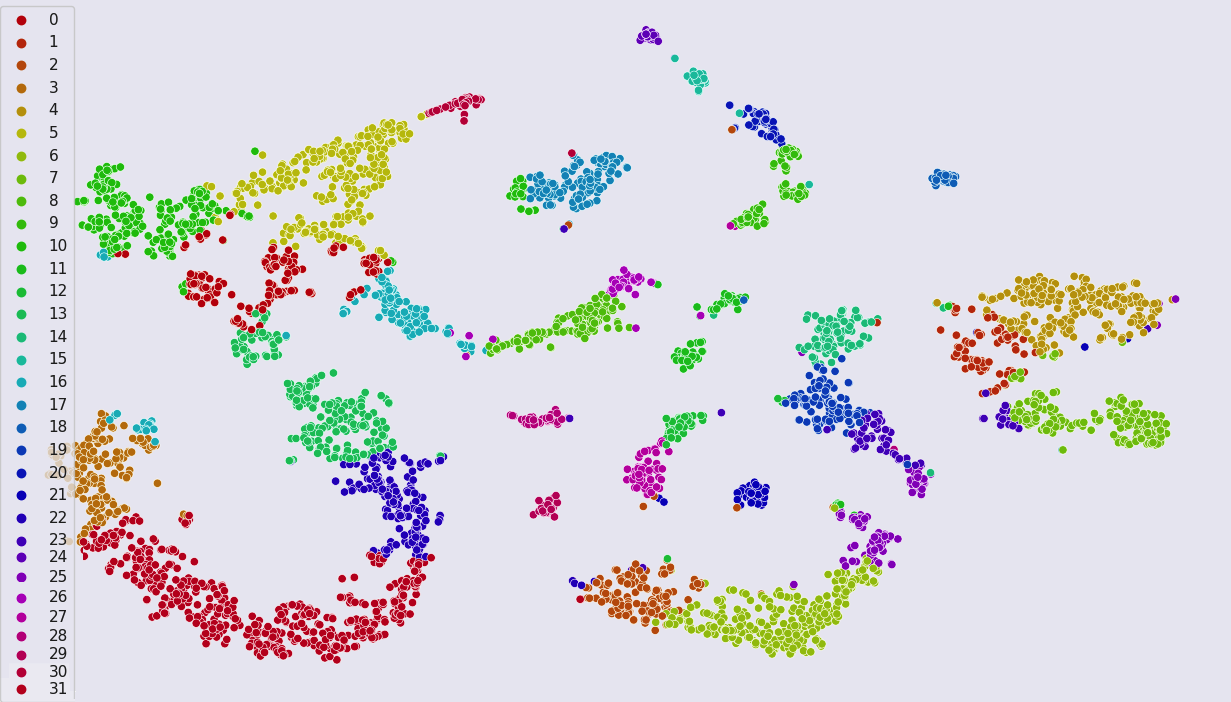}
}}
\caption{2D t-SNE projection of the testing set with color-coded labels based on (a) arXiv categories (green) and (b) K-means clustering.}
\label{fig:test_embedding_CLS}
\end{figure}
%\begin{figure}[t!]
%\centering\includegraphics[width=0.7\textwidth]{img/crop_test_tuned_cls_Embedding_mostfrequentclass_from2022.png}
%\caption{2D projection of FT-SciBERT (CLS) embeddings of testing set. Labelled samples belonging the same arXiv subject category are shown in green, while the others in grey. The graph reports texts from the 6 most frequent categories.}
%\label{fig:test_embedding_CLS}
%\end{figure}

%\begin{figure}[t!]
%\centering\includegraphics[width=0.6\textwidth]{img/tuned_SciBERT_cls_kmeans_k=32_t-SNE_from2022.png}
%\caption{K-Means results on the testing set for $N=32$.}
%%\label{fig:kmeans}
%\end{figure}

\paragraph{Unsupervised text categorization}
A qualitative evaluation of FT-SciBERT-CLS embedding can be observed in \ref{fig:test_embedding_CLS} (a). The 2D t-SNE projection of the testing set shows samples colored based on the 6 most frequent arXiv categories (green), while the grey points represent other categories. The clustering of green points indicates that the embedding captures arXiv category information. However, there is some overlap between embeddings of \texttt{hep-th} (\textit{High Energy Physics - Theory}) and \texttt{gr-qc} (\textit{General Relativity and Quantum Cosmology}) -- two related scientific field --, suggesting that the FT-SciBERT-CLS representation goes beyond arXiv categories.

\begin{figure}[t!]
\centering\includegraphics[width=1.0\textwidth]{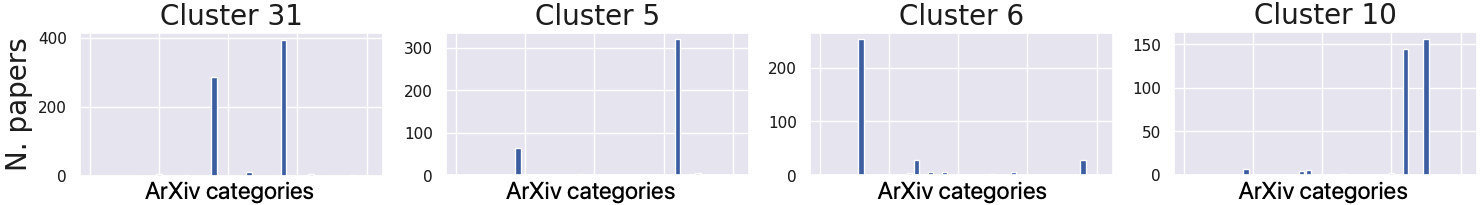}
\caption{Number of papers associated to arXiv cateogries in the 4 most populated K-Means clusters.}
\label{fig:clustercount}
\end{figure}

For the final evaluation of the categorization task, $N=32$ was chosen based on the results obtained on the validation set. %It is important to note that this optimal $N$ falls between the number of macro subject categories (15) and the number of subject categories (40), indicating that arXiv categories alone may not fully describe the abstract classes. 
The classification model achieved a Silhouette score of 0.4 on the testing set. Figure \ref{fig:test_embedding_CLS} (b) displays a t-SNE projection of the K-Means results, where the testing samples are color-coded based on their assigned K-Means class.
A more detailed analysis of the results on the testing set and the relationship between the identified and the arXiv categories can be seen in Figure \ref{fig:clustercount}. The figure presents four bar charts representing the four largest clusters. Each chart displays the distribution of papers across different arXiv categories within the cluster. Interestingly, most clusters exhibit one or a few dominant peaks, indicating a strong association with the arXiv category tags. This pattern is observed consistently across all K-means clusters.

A further examination was conducted to evaluate the top three most frequent subject categories within each cluster. Table \ref{table2} presents the list of the clusters sorted by decreasing cardinality, excluding arXiv categories with fewer than 10 samples. Results indicate that 66\% of clusters -- i.e. the identified categories -- correspond to a predominant arXiv category or multiple subject categories within the same macro-category. For example, cluster 8 contains samples associated with \textit{Astrophysics of Galaxies}, \textit{Astrophysics}, and \textit{Cosmology and Non-galactic Astrophysics}, all belonging to the macro-category of \textit{Astrophysics}. According to the results, 21\% of clusters have two main macro-categories, while only 7\% have three. Notably, the K-Means algorithm successfully classifies texts that share the same subject area but are labeled with different category tags in arXiv. For instance, the \texttt{stat.Th} and \texttt{math.ST} papers, which correspond to the same subject of \textit{Statistic Theory} are correctly clustered together in cluster 29.  Similarly, cluster 2 and cluster 6 group together abstracts labeled as \texttt{math-ph} and \texttt{mat.MP} respectively, both representing the subject of \textit{Mathematical Physics}. This demonstrates that the proposed approach effectively identifies and clusters abstracts based on their subject matter, overcoming the limitations of the original category labels and providing a more meaningful classification.

\begin{table}[t!]\small
\caption{Top 3 subject categories per cluster (number of papers per category in parenthesis).}
\label{table2}
\scalebox{0.6}{
\begin{tabular}{cccc}
 \toprule
 Cluster & 1st most frequent & 2nd most frequent & 3rd most frequent \\
  \midrule
31&General Relativity and Quantum Cosmology (394) & High Energy Physics - Theory (286)& -\\ 
 5&High Energy Physics - Phenomenology (320) & High Energy Physics - Experiment (63) & -\\ 
6&Quantum Physics (253)& Mathematical Physics (mat.MP) (28) & Mathematical Physics  (math-ph) (28)\\ 
10 & Nuclear Theory (156) & High Energy Physics - Phenomenology (145) &-\\ 
13 & High Energy Physics - Theory (163) & High Energy Physics - Phenomenology (145) & -\\ 
4 & Mesoscale and Nanoscale Physics (133)  & Materials Science (125) & Strongly Correlated Electrons (10)\\
3 & General Relativity and Quantum Cosmology (167) & Astrophysics (36) & High Energy Physics - Theory (23)\\ 
7 & Superconductivity (102) & Strongly Correlated Electrons (85) & Materials Science (11)\\ 
22 & High Energy Physics - Theory (162) & General Relativity and Quantum Cosmology (17) & High Energy Physics - Phenomenology (10)\\ 
0& High Energy Physics - Phenomenology (162) & Nuclear Theory (15) & -\\ 
16 & High Energy Physics - Phenomenology (83) & Astrophysics (74)& General Relativity and Quantum Cosmology (13)\\ 
8 & Astrophysics of Galaxies (54) & Astrophysics (39) & Cosmology and Non-galactic Astrophysics (34)\\ 
 2 & High Energy Physics  - Theory (44) & Mathematical Physics (math-ph) (37) & Mathematical Physics (mat.MP) (37)\\ 
 17 & Nuclear Theory (90) & Nuclear Experiment (36) & -\\ 
 25 & Quantum Physics (40) & Quantum Gases (25) & Atomic Physics (15)\\  
  19&Statistical Mechanics (78)& -&-\\ 
14&Soft Condensed Matter (77)& - & -\\ 
9&Geometric Topology (33)& Algebraic Geometry (28)& Differential Geometry (26)\\
 1 & Materials Science (72)& - &-\\ 
 11&Plasma Physics (33)& Fluid Dynamics (25)& Chemical Physics (10)\\
 23 & Strongly Correlated Electrons (24) & Statistical Mechanics (13)& -\\  
27 & Analysis of PDEs (42)& Dynamical Systems (15)& -\\ 
  12&Probability (42)& - &-\\
 20& Number Theory (24)& Algebraic Geometry (11)& -\\ 
 28 & High Energy Physics - Lattice (33) & High Energy Physics - Phenomenology (12) & - \\
 21 & Optics (35)& - & - \\
 30 & High Energy Physics - Experiment (42) & -& -\\
  26 & High Energy Astrophysical Phenomena (25) & Astrophysics (12)& -\\
 29 & Statistics Theory (stat.TH) (31) & Statistics Theory (math.ST) (31) & -\\
 18 & Computer Vision and Pattern Recognition (35) & -&-\\ 
 15 & Combinatorics (26)& - &-\\ 
 24&Logic in Computer Science (29) & -&-\\ 
 \bottomrule
 \end{tabular}
 }
\end{table}

\section{Conclusion}
This study introduces a novel approach for text categorization in scientific literature using pre-trained language models. By surpassing the limitations of traditional arXiv subject categories, this method enables more meaningful and accurate categorization of abstracts. The results demonstrate the effectiveness of the proposed approach in capturing the underlying subject matter, improving literature search and recommendation systems. This research contributes to the advancement of natural language processing techniques and addresses the need for more efficient tools in navigating scientific literature. Further studies will extend the proposed approach on other datasets from various scientific disciplines. 

%\begin{acknowledgments}
%  Thanks to the developers of ACM consolidated LaTeX styles
%  \url{https://github.com/borisveytsman/acmart} and to the developers
%  of Elsevier updated \LaTeX{} templates
%  \url{https://www.ctan.org/tex-archive/macros/latex/contrib/els-cas-templates}.  
%\end{acknowledgments}

%%
%% Define the bibliography file to be used
\bibliography{main}

\begin{thebibliography}{29}
\expandafter\ifx\csname natexlab\endcsname\relax\def\natexlab#1{#1}\fi
\providecommand{\url}[1]{\texttt{#1}}
\providecommand{\href}[2]{#2}
\providecommand{\path}[1]{#1}
\providecommand{\DOIprefix}{doi:}
\providecommand{\ArXivprefix}{arXiv:}
\providecommand{\URLprefix}{URL: }
\providecommand{\Pubmedprefix}{pmid:}
\providecommand{\doi}[1]{\href{http://dx.doi.org/#1}{\path{#1}}}
\providecommand{\Pubmed}[1]{\href{pmid:#1}{\path{#1}}}
\providecommand{\bibinfo}[2]{#2}
\ifx\xfnm\relax \def\xfnm[#1]{\unskip,\space#1}\fi
%Type = Misc
\bibitem[{Ginsparg(1991)}]{arxiv}
\bibinfo{author}{P.~Ginsparg}, \bibinfo{title}{arxiv}, \bibinfo{year}{1991}. \URLprefix \url{https://arxiv.org/}.
%Type = Article
\bibitem[{Qiu et~al.(2020)Qiu, Sun, Xu, Shao, Dai, and Huang}]{qiu2020pre}
\bibinfo{author}{X.~Qiu}, \bibinfo{author}{T.~Sun}, \bibinfo{author}{Y.~Xu}, \bibinfo{author}{Y.~Shao}, \bibinfo{author}{N.~Dai}, \bibinfo{author}{X.~Huang},
\newblock \bibinfo{title}{Pre-trained models for natural language processing: A survey},
\newblock \bibinfo{journal}{Science China Technological Sciences} \bibinfo{volume}{63} (\bibinfo{year}{2020}) \bibinfo{pages}{1872--1897}.
%Type = Article
\bibitem[{Bianchi et~al.(2020)Bianchi, Terragni, Hovy, Nozza, and Fersini}]{bianchi2020cross}
\bibinfo{author}{F.~Bianchi}, \bibinfo{author}{S.~Terragni}, \bibinfo{author}{D.~Hovy}, \bibinfo{author}{D.~Nozza}, \bibinfo{author}{E.~Fersini},
\newblock \bibinfo{title}{Cross-lingual contextualized topic models with zero-shot learning},
\newblock \bibinfo{journal}{arXiv preprint arXiv:2004.07737}  (\bibinfo{year}{2020}).
%Type = Inproceedings
\bibitem[{Ramage et~al.(2009)Ramage, Rosen, Chuang, Manning, and McFarland}]{ramage2009topic}
\bibinfo{author}{D.~Ramage}, \bibinfo{author}{E.~Rosen}, \bibinfo{author}{J.~Chuang}, \bibinfo{author}{C.~D. Manning}, \bibinfo{author}{D.~A. McFarland},
\newblock \bibinfo{title}{Topic modeling for the social sciences},
\newblock in: \bibinfo{booktitle}{NIPS 2009 workshop on applications for topic models: text and beyond}, volume~\bibinfo{volume}{5}, \bibinfo{year}{2009}, pp. \bibinfo{pages}{1--4}.
%Type = Article
\bibitem[{Grootendorst(2022)}]{grootendorst2022bertopic}
\bibinfo{author}{M.~Grootendorst},
\newblock \bibinfo{title}{Bertopic: Neural topic modeling with a class-based tf-idf procedure},
\newblock \bibinfo{journal}{arXiv preprint arXiv:2203.05794}  (\bibinfo{year}{2022}).
%Type = Article
\bibitem[{Yang et~al.(2019)Yang, Dai, Yang, Carbonell, Salakhutdinov, and Le}]{yang2019xlnet}
\bibinfo{author}{Z.~Yang}, \bibinfo{author}{Z.~Dai}, \bibinfo{author}{Y.~Yang}, \bibinfo{author}{J.~Carbonell}, \bibinfo{author}{R.~R. Salakhutdinov}, \bibinfo{author}{Q.~V. Le},
\newblock \bibinfo{title}{Xlnet: Generalized autoregressive pretraining for language understanding},
\newblock \bibinfo{journal}{Advances in neural information processing systems} \bibinfo{volume}{32} (\bibinfo{year}{2019}).
%Type = Inproceedings
\bibitem[{Sun et~al.(2019)Sun, Qiu, Xu, and Huang}]{sun2019fine}
\bibinfo{author}{C.~Sun}, \bibinfo{author}{X.~Qiu}, \bibinfo{author}{Y.~Xu}, \bibinfo{author}{X.~Huang},
\newblock \bibinfo{title}{How to fine-tune bert for text classification?},
\newblock in: \bibinfo{booktitle}{Chinese Computational Linguistics: 18th China National Conference, CCL 2019, Kunming, China, October 18--20, 2019, Proceedings 18}, \bibinfo{organization}{Springer}, \bibinfo{year}{2019}, pp. \bibinfo{pages}{194--206}.
%Type = Inproceedings
\bibitem[{Liu et~al.(2018)Liu, Huang, Gao, Wei, Tian, and Liu}]{liu2018task}
\bibinfo{author}{Q.~Liu}, \bibinfo{author}{H.-Y. Huang}, \bibinfo{author}{Y.~Gao}, \bibinfo{author}{X.~Wei}, \bibinfo{author}{Y.~Tian}, \bibinfo{author}{L.~Liu},
\newblock \bibinfo{title}{Task-oriented word embedding for text classification},
\newblock in: \bibinfo{booktitle}{Proceedings of the 27th international conference on computational linguistics}, \bibinfo{year}{2018}, pp. \bibinfo{pages}{2023--2032}.
%Type = Article
\bibitem[{Wang et~al.(2017)Wang, Mao, Wang, and Guo}]{wang2017knowledge}
\bibinfo{author}{Q.~Wang}, \bibinfo{author}{Z.~Mao}, \bibinfo{author}{B.~Wang}, \bibinfo{author}{L.~Guo},
\newblock \bibinfo{title}{Knowledge graph embedding: A survey of approaches and applications},
\newblock \bibinfo{journal}{IEEE Transactions on Knowledge and Data Engineering} \bibinfo{volume}{29} (\bibinfo{year}{2017}) \bibinfo{pages}{2724--2743}.
%Type = Article
\bibitem[{Gonzalez-Marquez et~al.(2023)Gonzalez-Marquez, Schmidt, Schmidt, Berens, and Kobak}]{gonzalez2023landscape}
\bibinfo{author}{R.~Gonzalez-Marquez}, \bibinfo{author}{L.~Schmidt}, \bibinfo{author}{B.~M. Schmidt}, \bibinfo{author}{P.~Berens}, \bibinfo{author}{D.~Kobak},
\newblock \bibinfo{title}{The landscape of biomedical research},
\newblock \bibinfo{journal}{bioRxiv}  (\bibinfo{year}{2023}) \bibinfo{pages}{2023--04}.
%Type = Inproceedings
\bibitem[{Lumbanraja et~al.(2021)Lumbanraja, Fitri, Junaidi, Prabowo et~al.}]{lumbanraja2021abstract}
\bibinfo{author}{F.~R. Lumbanraja}, \bibinfo{author}{E.~Fitri}, \bibinfo{author}{A.~Junaidi}, \bibinfo{author}{R.~Prabowo}, et~al.,
\newblock \bibinfo{title}{Abstract classification using support vector machine algorithm (case study: abstract in a computer science journal)},
\newblock in: \bibinfo{booktitle}{Journal of Physics: Conference Series}, volume \bibinfo{volume}{1751}, \bibinfo{organization}{IOP Publishing}, \bibinfo{year}{2021}, p. \bibinfo{pages}{012042}.
%Type = Article
\bibitem[{Beltagy et~al.(2019)Beltagy, Lo, and Cohan}]{beltagy2019SciBERT}
\bibinfo{author}{I.~Beltagy}, \bibinfo{author}{K.~Lo}, \bibinfo{author}{A.~Cohan},
\newblock \bibinfo{title}{Scibert: A pretrained language model for scientific text},
\newblock \bibinfo{journal}{arXiv preprint arXiv:1903.10676}  (\bibinfo{year}{2019}).
%Type = Misc
\bibitem[{MES(1960)}]{MESH}
\bibinfo{title}{Medical subject headings (mesh)}, \bibinfo{year}{1960}. \URLprefix \url{https://www.nlm.nih.gov/mesh/meshhome.html}.
%Type = Misc
\bibitem[{Phy(2016)}]{PhySH}
\bibinfo{title}{Physics subject headings (physh)}, \bibinfo{year}{2016}. \URLprefix \url{https://physh.aps.org}.
%Type = Misc
\bibitem[{STW(1998)}]{STW}
\bibinfo{title}{Stw thesaurus for economics}, \bibinfo{year}{1998}. \URLprefix \url{http://zbw.eu/stw/version/latest/about}.
%Type = Inproceedings
\bibitem[{Mai et~al.(2018)Mai, Galke, and Scherp}]{mai2018using}
\bibinfo{author}{F.~Mai}, \bibinfo{author}{L.~Galke}, \bibinfo{author}{A.~Scherp},
\newblock \bibinfo{title}{Using deep learning for title-based semantic subject indexing to reach competitive performance to full-text},
\newblock in: \bibinfo{booktitle}{Proceedings of the 18th ACM/IEEE on joint conference on digital libraries}, \bibinfo{year}{2018}, pp. \bibinfo{pages}{169--178}.
%Type = Inproceedings
\bibitem[{Galke et~al.(2017)Galke, Mai, Schelten, Brunsch, and Scherp}]{galke2017using}
\bibinfo{author}{L.~Galke}, \bibinfo{author}{F.~Mai}, \bibinfo{author}{A.~Schelten}, \bibinfo{author}{D.~Brunsch}, \bibinfo{author}{A.~Scherp},
\newblock \bibinfo{title}{Using titles vs. full-text as source for automated semantic document annotation},
\newblock in: \bibinfo{booktitle}{Proceedings of the Knowledge Capture Conference}, \bibinfo{year}{2017}, pp. \bibinfo{pages}{1--4}.
%Type = Inproceedings
\bibitem[{Nishioka and Scherp(2016)}]{nishioka2016profiling}
\bibinfo{author}{C.~Nishioka}, \bibinfo{author}{A.~Scherp},
\newblock \bibinfo{title}{Profiling vs. time vs. content: What does matter for top-k publication recommendation based on twitter profiles?},
\newblock in: \bibinfo{booktitle}{Proceedings of the 16th ACM/IEEE-CS on joint conference on digital libraries}, \bibinfo{year}{2016}, pp. \bibinfo{pages}{171--180}.
%Type = Article
\bibitem[{Kandimalla et~al.(2021)Kandimalla, Rohatgi, Wu, and Giles}]{kandimalla2021large}
\bibinfo{author}{B.~Kandimalla}, \bibinfo{author}{S.~Rohatgi}, \bibinfo{author}{J.~Wu}, \bibinfo{author}{C.~L. Giles},
\newblock \bibinfo{title}{Large scale subject category classification of scholarly papers with deep attentive neural networks},
\newblock \bibinfo{journal}{Frontiers in research metrics and analytics} \bibinfo{volume}{5} (\bibinfo{year}{2021}) \bibinfo{pages}{600382}.
%Type = Inproceedings
\bibitem[{Osborne and Motta(2012)}]{osborne2012mining}
\bibinfo{author}{F.~Osborne}, \bibinfo{author}{E.~Motta},
\newblock \bibinfo{title}{Mining semantic relations between research areas},
\newblock in: \bibinfo{booktitle}{The Semantic Web--ISWC 2012: 11th International Semantic Web Conference, Boston, MA, USA, November 11-15, 2012, Proceedings, Part I 11}, \bibinfo{organization}{Springer}, \bibinfo{year}{2012}, pp. \bibinfo{pages}{410--426}.
%Type = Inproceedings
\bibitem[{Er{\'e}t{\'e}o et~al.(2011)Er{\'e}t{\'e}o, Gandon, and Buffa}]{ereteo2011semtagp}
\bibinfo{author}{G.~Er{\'e}t{\'e}o}, \bibinfo{author}{F.~Gandon}, \bibinfo{author}{M.~Buffa},
\newblock \bibinfo{title}{Semtagp: semantic community detection in folksonomies},
\newblock in: \bibinfo{booktitle}{2011 IEEE/WIC/ACM International Conferences on Web Intelligence and Intelligent Agent Technology}, volume~\bibinfo{volume}{1}, \bibinfo{organization}{IEEE}, \bibinfo{year}{2011}, pp. \bibinfo{pages}{324--331}.
%Type = Inproceedings
\bibitem[{Salatino et~al.(2019)Salatino, Osborne, Thanapalasingam, and Motta}]{salatino2019cso}
\bibinfo{author}{A.~A. Salatino}, \bibinfo{author}{F.~Osborne}, \bibinfo{author}{T.~Thanapalasingam}, \bibinfo{author}{E.~Motta},
\newblock \bibinfo{title}{The cso classifier: Ontology-driven detection of research topics in scholarly articles},
\newblock in: \bibinfo{booktitle}{Digital Libraries for Open Knowledge: 23rd International Conference on Theory and Practice of Digital Libraries, TPDL 2019, Oslo, Norway, September 9-12, 2019, Proceedings 23}, \bibinfo{organization}{Springer}, \bibinfo{year}{2019}, pp. \bibinfo{pages}{296--311}.
%Type = Inproceedings
\bibitem[{Hitha and Kiran(2021)}]{hitha2021topic}
\bibinfo{author}{K.~Hitha}, \bibinfo{author}{V.~Kiran},
\newblock \bibinfo{title}{Topic recognition and correlation analysis of articles in computer science},
\newblock in: \bibinfo{booktitle}{2021 Fifth International Conference on I-SMAC (IoT in Social, Mobile, Analytics and Cloud)(I-SMAC)}, \bibinfo{organization}{IEEE}, \bibinfo{year}{2021}, pp. \bibinfo{pages}{1115--1118}.
%Type = Inproceedings
\bibitem[{Al-Shareef et~al.(2022)Al-Shareef, Alharbi, Alharbi, Almfarriji, Alsharif, Alharthi, and Althaqafi}]{al2022investigating}
\bibinfo{author}{S.~Al-Shareef}, \bibinfo{author}{R.~Alharbi}, \bibinfo{author}{R.~Alharbi}, \bibinfo{author}{R.~Almfarriji}, \bibinfo{author}{M.~Alsharif}, \bibinfo{author}{R.~Alharthi}, \bibinfo{author}{L.~Althaqafi},
\newblock \bibinfo{title}{Investigating community detection in arabic scholarly network using ontology-based semantic expansion},
\newblock in: \bibinfo{booktitle}{2022 IEEE/ACM International Conference on Advances in Social Networks Analysis and Mining (ASONAM)}, \bibinfo{organization}{IEEE}, \bibinfo{year}{2022}, pp. \bibinfo{pages}{96--103}.
%Type = Article
\bibitem[{Sharma and Kumar(2023)}]{sharma2023machine}
\bibinfo{author}{A.~Sharma}, \bibinfo{author}{S.~Kumar},
\newblock \bibinfo{title}{Machine learning and ontology-based novel semantic document indexing for information retrieval},
\newblock \bibinfo{journal}{Computers \& Industrial Engineering} \bibinfo{volume}{176} (\bibinfo{year}{2023}) \bibinfo{pages}{108940}.
%Type = Misc
\bibitem[{Library(2019)}]{arxiv_dataset}
\bibinfo{author}{C.~U. Library}, \bibinfo{year}{2019}. \URLprefix \url{https://www.kaggle.com/datasets/Cornell-University/arxiv}.
%Type = Unpublished
\bibitem[{Honnibal and Montani(2017)}]{spacy2}
\bibinfo{author}{M.~Honnibal}, \bibinfo{author}{I.~Montani}, \bibinfo{title}{{spaCy 2}: Natural language understanding with {B}loom embeddings, convolutional neural networks and incremental parsing}, \bibinfo{year}{2017}. \bibinfo{note}{To appear}.
%Type = Inproceedings
\bibitem[{Wolf et~al.(2020)Wolf, Debut, Sanh, Chaumond, Delangue, Moi, Cistac, Rault, Louf, Funtowicz, Davison, Shleifer, von Platen, Ma, Jernite, Plu, Xu, Scao, Gugger, Drame, Lhoest, and Rush}]{wolf-etal-2020-transformers}
\bibinfo{author}{T.~Wolf}, \bibinfo{author}{L.~Debut}, \bibinfo{author}{V.~Sanh}, \bibinfo{author}{J.~Chaumond}, \bibinfo{author}{C.~Delangue}, \bibinfo{author}{A.~Moi}, \bibinfo{author}{P.~Cistac}, \bibinfo{author}{T.~Rault}, \bibinfo{author}{R.~Louf}, \bibinfo{author}{M.~Funtowicz}, \bibinfo{author}{J.~Davison}, \bibinfo{author}{S.~Shleifer}, \bibinfo{author}{P.~von Platen}, \bibinfo{author}{C.~Ma}, \bibinfo{author}{Y.~Jernite}, \bibinfo{author}{J.~Plu}, \bibinfo{author}{C.~Xu}, \bibinfo{author}{T.~L. Scao}, \bibinfo{author}{S.~Gugger}, \bibinfo{author}{M.~Drame}, \bibinfo{author}{Q.~Lhoest}, \bibinfo{author}{A.~M. Rush},
\newblock \bibinfo{title}{Transformers: State-of-the-art natural language processing},
\newblock in: \bibinfo{booktitle}{Proceedings of the 2020 Conference on Empirical Methods in Natural Language Processing: System Demonstrations}, \bibinfo{publisher}{Association for Computational Linguistics}, \bibinfo{address}{Online}, \bibinfo{year}{2020}, pp. \bibinfo{pages}{38--45}. \URLprefix \url{https://www.aclweb.org/anthology/2020.emnlp-demos.6}.
%Type = Article
\bibitem[{van~der Maaten and Hinton(2008)}]{vandermaaten08a}
\bibinfo{author}{L.~van~der Maaten}, \bibinfo{author}{G.~Hinton},
\newblock \bibinfo{title}{Visualizing data using t-sne},
\newblock \bibinfo{journal}{Journal of Machine Learning Research} \bibinfo{volume}{9} (\bibinfo{year}{2008}) \bibinfo{pages}{2579--2605}. \URLprefix \url{http://jmlr.org/papers/v9/vandermaaten08a.html}.

\end{thebibliography}

%%
%% If your work has an Supplementary Materials, this is the place to put it.

%\begin{figure}[t!]
%\centering\includegraphics[width=0.75\textwidth]{img/test_tuned_Embedding_mostfrequentclass_from2022.png}
%\caption{2D projection of FT-SciBERT-T embeddings of testing set. Labelled samples belonging the same arXiv subject category are shown in green, while the others in grey. The graph reports texts from the 6 most frequent categories.}
%\label{fig:test_embedding_T}
%\end{figure}

\end{document}